\documentclass[11pt,a4paper]{article}
\usepackage[hyperref]{acl2020}
\usepackage{times}
\usepackage{latexsym}
\usepackage{xcolor}
\usepackage[ruled,vlined]{algorithm2e}

\usepackage{booktabs}

\definecolor{c1}{HTML}{0072B2}%
\definecolor{c2}{HTML}{D55E00}%
\definecolor{c3}{HTML}{009E73}%
\definecolor{c4}{HTML}{56B4E9}%
\definecolor{c5}{HTML}{CC79A7}%
\definecolor{c6}{HTML}{E69F00}%
\definecolor{c7}{HTML}{844E4D}%
\definecolor{c8}{HTML}{2D512A}%

\newcommand\one[1]{\textcolor{c1}{\textbf{#1}}}
\newcommand\two[1]{\textcolor{c2}{\textbf{#1}}}
\newcommand\three[1]{\textcolor{c3}{\textbf{#1}}}
\newcommand\four[1]{\textcolor{c4}{\textbf{#1}}}
\newcommand\five[1]{\textcolor{c5}{\textbf{#1}}}
\newcommand\six[1]{\textcolor{c6}{\textbf{#1}}}
\newcommand\seven[1]{\textcolor{c7}{\textbf{#1}}}
\newcommand\eight[1]{\textcolor{c8}{\textbf{#1}}}

\usepackage{microtype}

\aclfinalcopy %
\def\aclpaperid{1852} %

\usepackage{graphicx}
\usepackage{xcolor}
\usepackage{url}
\usepackage{amsmath}
\usepackage{siunitx}

\usepackage{multirow}
\usepackage{subfigure}

\newcommand{\hotel}{\mbox{\sc Hotel}}
\newcommand{\yelp}{\mbox{\sc Yelp}}
\newcommand{\NAME}{\mbox{\sc OpinionDigest}}

\newcommand\blfootnote[1]{%
  \begingroup
  \renewcommand\thefootnote{}\footnote{#1}%
  \addtocounter{footnote}{-1}%
  \endgroup
}

\title{\NAME: A Simple Framework for Opinion Summarization}

\author{Yoshihiko Suhara$^{*1}$%
\quad Xiaolan Wang$^{*1}$
\quad Stefanos Angelidis$^2$
\quad Wang-Chiew Tan$^1$ \\
  $^1$Megagon Labs \quad
  $^2$University of Edinburgh \\
  \texttt{\{yoshi,xiaolan,wangchiew\}@megagon.ai}\quad \texttt{s.angelidis@ed.ac.uk}
}

\date{}

\begin{document}
\maketitle
\begin{abstract}
We present \NAME, an abstractive opinion summarization framework, which does not rely on gold-standard summaries for training. The framework uses an Aspect-based Sentiment Analysis model to extract opinion phrases from reviews, and trains a Transformer model to reconstruct the original reviews from these extractions.
At summarization time, we merge extractions from multiple reviews and select the most popular ones. The selected opinions are used as input to the trained Transformer model, which verbalizes them into an opinion summary.
\NAME{} can also generate customized summaries, tailored to specific user needs, by filtering the selected opinions according to their aspect and/or sentiment.
Automatic evaluation on {\sc Yelp} data shows that our framework outperforms competitive
baselines. Human studies on two corpora verify that \NAME{} produces informative summaries and shows promising customization capabilities\blfootnote{\hspace{-1.5mm}$^{*}$Equal contribution.}\footnote{Our code is available at \url{https://github.com/megagonlabs/opiniondigest}.}.
\end{abstract}

\section{Introduction}
The summarization of opinions in customer reviews has received significant attention in the Data Mining and Natural Language Processing communities. Early efforts \citep{Hu:2004:CustomerReviews} focused on producing \textsl{structured} summaries which numerically aggregate the customers' satisfaction about an item across multiple aspects, and often included representative review sentences as evidence. Considerable research has recently shifted towards textual opinion summaries, fueled by the increasing success of neural summarization methods \cite{cheng-lapata-2016-neural,Paulus:2017:DeepReinforced,See:2017:PointerGenerator,liu-lapata-2019-hierarchical,Isonuma:2019:Unsupervised}.

Opinion summaries can be extractive, i.e., created by selecting a subset of salient sentences from the input reviews, or abstractive, where summaries are generated from scratch. Extractive approaches produce well-formed text, but selecting the sentences which approximate the most popular opinions in the input is challenging. \citet{Angelidis:2018:MATE} used sentiment and aspect predictions as a proxy for identifying opinion-rich segments. Abstractive methods  \cite{Chu:2019:MeanSum,Brainskas:2019:Unsupervised}, like the one presented in this paper, attempt to model the prevalent opinions in the input and generate text that articulates them.

Opinion summarization can rarely rely on gold-standard summaries for training (see \citet{amplayo2019informative} for a supervised approach). Recent work has utilized end-to-end unsupervised architectures, based on auto-encoders \cite{Chu:2019:MeanSum,Brainskas:2019:Unsupervised}, where an aggregated representation of the input reviews is fed to a decoder, trained via reconstruction loss to produce review-like summaries. Similarly to their work, we assume that review-like generation is appropriate for opinion summarization. However, we explicitly deal with opinion \textsl{popularity}, which we believe is crucial for multi-review opinion summarization. Additionally, our work is novel in its ability to explicitly control the sentiment and aspects of selected opinions. The aggregation of input reviews is no longer treated as a black box, thus allowing for controllable summarization.

\begin{figure*}[t]
    \centering
    \includegraphics[width=\textwidth]{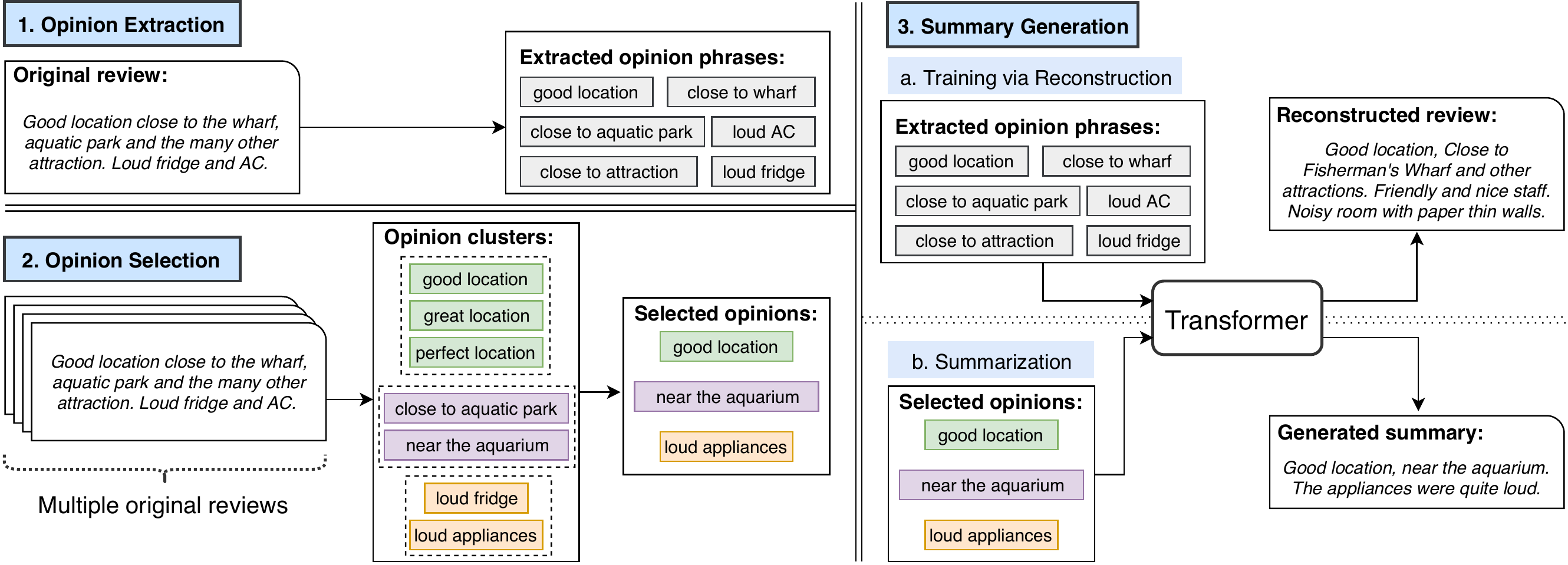}
    \caption{Overview of the \NAME{} framework.}
    \label{fig:overview}
    \vspace{-1em}
\end{figure*}

Specifically, we take a step towards more interpretable and controllable opinion aggregation, as we replace the end-to-end architectures of previous work with a pipeline framework. Our method has three components: a) a pre-trained opinion extractor, which identifies opinion phrases in reviews; b) a simple and controllable opinion selector, which merges, ranks, and --optionally-- filters the extracted opinions; and c) a generator model, which is trained to reconstruct reviews from their extracted opinion phrases and can then generate opinion summaries based on the selected opinions.

We describe our framework in Section \ref{sec:framework} and present two types of experiments in Section \ref{sec:eval}: A quantitative comparison against established summarization techniques on the \textsc{Yelp} 
summarization corpus \citep{Chu:2019:MeanSum}; and two user studies, validating the automatic results and our method's ability for controllable summarization.

\section{\NAME{} Framework}
\label{sec:framework}

Let~$D$ denote a dataset of customer reviews on individual entities $\{e_1, e_2, \dots, e_{|D|}\}$ from a single domain, e.g.,~restaurants or hotels. For every entity~$e$, we define a review set \mbox{$R_e = \{r_i\}_{i=1}^{|R_e|}$}, where each review is a sequence of words \mbox{$r = (w_{1}, \dots, w_{n})$}.

Within a review, we define a single opinion phrase, $o = (w_{o1}, \dots w_{om})$, as a subsequence of tokens that expresses the attitude of the reviewer towards a specific aspect of the entity\footnote{Words that form an opinion may not be contiguous in the review. Additionally, a word can be part of multiple opinions.}. Formally, we define the \textsl{opinion set} of $r$ as \mbox{$O_r = \{(o_i, pol_i, a_i)\}_{i=1}^{|O_r|}$}, where $pol_i$ is the sentiment polarity of the $i$-th phrase (\textsl{positive}, \textsl{neutral}, or \textsl{negative}) and $a_i$ is the aspect category it discusses (e.g., a hotel's \textsl{service}, or \textsl{cleanliness}).

For each entity $e$, our task is to abstractively generate a summary $s_e$ of the most salient opinions expressed in reviews $R_e$. Contrary to previous abstractive methods \citep{Chu:2019:MeanSum,Brainskas:2019:Unsupervised}, which never explicitly deal with opinion phrases, we put the opinion sets of reviews at the core of our framework, as described in the following sections and illustrated in Figure \ref{fig:overview}.

\subsection{Opinion Extraction}
Extracting opinion phrases from reviews has been studied for years under the Aspect-based Sentiment Analysis (ABSA) task~\cite{hu2004mining, luo2019doer, dai2019neural, Li:2019:Opine}.%

We follow existing approaches to obtain an opinion set $O_r$ for every review in our corpus\footnote{Our framework is flexible with respect to the choice of opinion extraction models.}.

Specifically, we used a pre-trained tagging model~\cite{Miao:2020:Snippext}
to extract opinion phrases, their polarity, and aspect categories.
Step 1 (top-left) of Figure~\ref{fig:overview} shows a set of opinions extracted from a full review.

\subsection{Opinion Selection}
\label{sec:selection}
Given the set or reviews $R_e = \{r_1, r_2, \dots\}$ for an entity $e$, we define the \textsl{entity}'s opinion set as $O_e = \{O_{r_1} \cup O_{r_2} \cup \dots\}$. Summarizing the opinions about entity $e$ relies on selecting the most salient opinions $S_e \subset O_e$. As a departure from previous work, we explicitly select the opinion phrases that will form the basis for summarization, in the following steps.

\paragraph{Opinion Merging:} To avoid selecting redundant opinions in $S_e$, we apply a greedy algorithm to merge similar opinions into clusters \mbox{$\mathbf{C}=\{C_1, C_2, ...\}$}: given an opinion set $O_e$, we start with an empty $\mathbf{C}$, and iterate through every opinion in $O_e$. For each opinion, $(o_i, pol_i, a_i)$, we further iterate through every existing cluster in random order. The opinion is added to the first cluster $C$ which satisfies the following criterion, or to a newly created cluster otherwise: 
{
\setlength{\abovedisplayskip}{8pt}%
\setlength{\belowdisplayskip}{8pt}%
\[
\forall (o_j, pol_j, a_j) \in C, \; cos(\mathrm{v}_i, \mathrm{v}_j) \geq \theta, 
\]
}%
\noindent where $\mathrm{v}_i$ and $\mathrm{v}_j$ are the average word embedding of opinion phrase $o_i$ and $o_j$ respectively, $cos(\cdot,\cdot)$ is the cosine similarity, and $\theta \in (0, 1]$ is a hyper-parameter. For each opinion cluster $\{C_1, C_2, \dots\}$, we define its representative opinion $Repr(C_i)$, which is the opinion phrase closest to its centroid. %

\paragraph{Opinion Ranking:} %
We assume that larger clusters contain opinions which are popular among reviews and, therefore, should have higher priority to be included in $S_e$. We use the representative opinions of the top-$k$ largest clusters, as selected opinions $S_e$. The Opinion Merging and Ranking steps are demonstrated in Step 2 (bottom-left) of Figure~\ref{fig:overview}, where the top-3 opinion clusters are shown and their representative opinions are selected.

\paragraph{Opinion Filtering (optional):} We can further control the selection by filtering opinions based on their predicted aspect category or sentiment polarity. For example, we may only allow opinions where $a_i = ``cleanliness"$.

\subsection{Summary Generation}

Our goal is to generate a natural language summary which articulates $S_e$, the set of selected 
opinions. To achieve this, we need a natural language generation (NLG) model which takes a set of opinion phrases as input and produces a fluent, \mbox{review-like} summary as output. Because we cannot rely on gold-standard summaries for training, we train an NLG model that encodes the extracted opinion phrases of a \textsl{single} review and then attempts to reconstruct the review's full text. Then, the trained model can be used to generate summaries.

\paragraph{Training via Review Reconstruction:} Having extracted $O_r$ for every review $r$ in a corpus, we construct training examples $\{T(O_r), r\}$, where $T(O_r)$ is a textualization of the review's opinion set, where all opinion phrases are concatenated in their original order, using a special token \texttt{[SEP]}. For example:
{
\setlength{\abovedisplayskip}{7pt}%
\setlength{\belowdisplayskip}{7pt}%
\begin{gather*}
O_r = \{\textrm{\textit{very comfy bed}, \textit{clean bath}}\} \\
T(O_r) = \textrm{``very comfy bed \texttt{[SEP]} clean bath''}
\end{gather*}
}%
\noindent The $\{T(O_r), r\}$ pairs are used to train a Transformer model~\cite{Vaswani:2017:Transformer}\footnote{Our framework is flexible w.r.t. the choice of the model. Using a pre-trained language model is part of future work.} to reconstruct review text from extracted opinions, as shown in Step 3a (top-right) of Figure \ref{fig:overview}.

\paragraph{Summarization:} At summarization time, we use the textualization of the selected opinions, $T(S_e)$, as input to the trained Transformer, which generates a natural language summary $s_e$ as output (Figure \ref{fig:overview}, Step 3b). We order the selected opinions by frequency (i.e., their respective cluster's size), but any desired ordering may be used.

\section{Evaluation}
\label{sec:eval}

\subsection{Datasets}
We used two review datasets for evaluation. The public \yelp{} corpus of restaurant reviews, previously used by \citet{Chu:2019:MeanSum}. We used a different snapshot of the data, filtered to the same specifications as the original paper, resulting in 624K training reviews. We used the same gold-standard summaries for 200 restaurants as used in \citet{Chu:2019:MeanSum}.

We also used \hotel{}, a private hotel review dataset that consists of 688K reviews for 284 hotels collected from multiple hotel booking websites. There are no gold-standard summaries for this dataset, so systems were evaluated by humans.

\subsection{Baselines}

\noindent\textbf{LexRank \normalfont{\cite{Erkan:2004:LexRank}}:} A popular unsupervised extractive summarization method. It selects sentences based on centrality scores calculated on a graph-based sentence similarity.

\smallskip

\noindent\textbf{MeanSum \normalfont{\cite{Chu:2019:MeanSum}}:} An unsupervised multi-document abstractive summarizer that minimizes a combination of reconstruction and vector similarity losses. We only applied MeanSum to \yelp{},  due to its requirement for a pre-trained language model, which was not available for \hotel. %

\smallskip

\noindent\textbf{Best Review / Worst Review \normalfont{\cite{Chu:2019:MeanSum}}:} A single review that has the highest/lowest average word overlap with the input reviews.

\subsection{Experimental Settings}

\begin{table}[t]
\small
    \centering
    \begin{tabular}{lcccc}\hline
    Method & R1 & R2 & RL \\\hline
    Best Review & 27.97 & 3.46 & 15.29 \\ %
    Worst Review & 16.91 & 1.66 & 11.11 \\ %
    LexRank & 24.62 & 3.03 & 14.43 \\
    MeanSum & 27.86 & 3.95 & 16.56 \\
    \NAME & {\bf 29.30} & \textbf{5.77} & {\bf 18.56} \\\hline %
    \end{tabular}
    \caption{Summarization results on \yelp{} with ROUGE.}
    \label{tab:results_yelp}
    \normalsize
   \vspace{-1.5em}
\end{table}

For opinion extraction, the ABSA models are trained with 1.3K labeled review sentences for \yelp{} and 2.4K for \hotel{}. For opinion merging, we used pre-trained word embeddings ({\tt glove.6B.300d}), $\theta = 0.8$, and selected the top-$k$ ($k=15$) most popular opinion clusters.

We trained a Transformer with the original architecture~\cite{Vaswani:2017:Transformer}. 
We used SGD with an initial learning rate of 0.1, a momentum of $\beta=0.1$, and a decay of $\gamma=0.1$ for 5 epochs with a batch size of 8.
For decoding, we used Beam Search with a beam size of 5, a length penalty of 0.6,
3-gram blocking~\cite{Paulus:2017:DeepReinforced}, and a maximum generation length of 60. We tuned hyper-parameters on the dev set, and our system appears robust to their setting (see Appendix~\ref{app:sensitivity}).

We performed automatic evaluation on the \yelp{} dataset with ROUGE-1 (R1), ROUGE-2 (R2), and ROUGE-L (RL)~\cite{Lin:2004:ROUGE} scores based on the 200 reference summaries~\cite{Chu:2019:MeanSum}.
We also conducted user studies on both \yelp{} and \hotel{} datasets to further understand 
the performance of different models.

\begin{table}[t]
\small
    \centering
    \begin{tabular}{lccc}\hline
    Method & I-score & C-score & R-score \\\hline
    LexRank & -35.4 & -32.1 & -13.5 \\
    MeanSum & 14.2 & 4.9 & {\bf 9.0} \\
    \NAME & {\bf 21.2} & {\bf 27.2} & 4.4 \\\hline
    \end{tabular}
    \\(a) \yelp \\
    \vspace{0.6em}
    \begin{tabular}{lccc}\hline
    Method & I-score & C-score & R-score \\\hline
    LexRank & -5.8 & -3.2 & -0.5 \\
    Best Review & -4.0 & -10.7 & {\bf 17.0} \\
    \NAME & {\bf 9.8} & {\bf 13.8} & -16.5 \\\hline
    \end{tabular}
    \\(b) \hotel 
    \caption{Best-Worst Scaling human evaluation.}
    \label{tab:human_task1_all}
    \normalsize
\end{table}

\begin{table}[t]
\small
    \centering
    \begin{tabular}{lccc}\hline
               & Fully ($\uparrow$) & Partially ($\uparrow$) & No ($\downarrow$) \\\hline
    MeanSum & 23.25 \% & 42.57 \% & 34.18 \% \\
    \NAME & {\bf 29.77 \%} & {\bf 47.91 \%} & {\bf 22.32 \%} \\\hline
    \end{tabular}
    \caption{Human evaluation results on content support.}
    \label{tab:human_task2}
    \normalsize
    \vspace{-5mm}
\end{table}

\begin{table}[t]
\small
    \centering
    \begin{tabular}{lccc}\hline
    \multicolumn{4}{c}{\textsl{Does the summary discuss the specified aspect:}}\\
    & Exclusively & Partially & Not \\\hline
    \hotel & 46.63 \% & 43.09 \% & 10.28 \% \\ \hline
    \end{tabular}
    \vspace{-2mm}
    \caption{User study on aspect-specific summaries.}
    \label{tab:human_task3}
    \vspace{-0.5cm}
    \normalsize
\end{table}

\begin{table*}[th]
\footnotesize
\begin{tabular}{
>{\centering\arraybackslash}m{1.7cm} | 
m{6.4cm} | 
m{6.6cm}}\toprule
Asp/Pol/N & \multicolumn{1}{c|}{Input opinions} & \multicolumn{1}{c}{Summary} \\\midrule
All/All/8 & \one{central location} [SEP] \underline{lovely hotel} [SEP] {\it recommend room} [SEP] \two{good breakfast} [SEP]  \one{very nice location} [SEP]  \three{very dedicated staff} [SEP]  \four{walking distance to coffee shops} [SEP] \six{perfect hotel} [SEP] \five{small bathroom} [SEP] {\it unkind personnel} 
& 
\one{This hotel is in a perfect location}, \four{walking distance to a lot of shops and restaurants}. \five{The room was small} but it was clean and comfortable.  \three{The staff were very nice and helpful.} \two{The breakfast was good} and \one{the location was very central}. I would \six{recommend this hotel} and would stay there again if visiting San Francisco.  \\\hline
All/All/128 & \one{great location} [SEP] \two{very friendly staff} [SEP] \underline{great place} [SEP] \three{walkable block from union square} [SEP] \five{very clean room} [SEP] \seven{really soft bed} [SEP] \eight{quite spacious room} [SEP] \six{comfortable view} [SEP] \four{nice rooms} [SEP] \underline{Only 20 minutes from airport} & 
\two{The staff were very friendly and helpful}. \eight{The rooms were quite spacious}. \seven{The bed was very comfortable} and \five{the room was clean}. \one{The location of the hotel is great} - \three{only a few minutes from union square}. \four{The room was really nice} and \six{the view was great}. \\\midrule
All/Positive/8 & \one{pretty good food} [SEP]  \four{good chicken curry} [SEP] \two{decent indian food} [SEP] {\it little hole in the} [SEP] \three{good variety of spices} [SEP] \underline{inexpensive food} [SEP] \four{yummy chicken tikka masala} [SEP] \two{good indian place} [SEP] \underline{stale food}
& \two{This is a good little indian food place}. \one{The food is good} and \four{the chicken tikka masala is pretty good}. \three{They have a good variety of indian food}.The chicken biryani is good but the chicken is a little dry. \\\hline
All/Negative/8 & \one{worst place} [SEP] \two{rude rude rude employees} [SEP] \underline{missing napkins} [SEP] {\it always rude order} [SEP] \underline{forgot silverware} [SEP] \three{cold food} [SEP] \four{dirty place} [SEP] \three{cold boneless wings} [SEP] \two{sucked service} [SEP] \two{blah waitress} [SEP] \two{horrible service} [SEP] \five{mediocre food}
& \one{This place is disgusting}. \five{The food is mediocre at best}. \two{The service is always rude} and \four{the place is dirty} and \two{the employees are rude and rude}. \three{The food has been cold} and \two{the service has been horrible}. The worst wings I've ever had in my entire life. The place has gone downhill since the last two times. \\\hline
Staff/All/8 & \one{sweetest staff} [SEP] \two{super service} [SEP] \two{great service} [SEP] \one{incredibly polite and professional staff} &
I have been coming here for years and have never had a bad experience. \one{The staff is super professional} and \two{the service is great}. \one{The staff are super professional as well}. I would highly recommend this place to anyone. \\\hline
Food/All/8 & \one{good food} [SEP] \two{great chinese food} [SEP] \underline{fresh food} [SEP] \three{favorite orange chicken} [SEP] \four{like shrimp fried rice} [SEP] \five{good portions} [SEP] \two{best chinese food} [SEP] {\it were really shrimp vegetable} [SEP] \underline{best orange beef} [SEP] {\it really good though flavor}
& \two{This is my favorite Chinese food} in the area. \one{The food is really good} and \five{the portions are great}. \three{I really like the orange chicken} and the crab puffs are the best I've had in a long time. \one{The food here is really good}. \four{The shrimp fried rice is really good}, and the rice is the best. \\\bottomrule
\end{tabular}
\normalsize
\caption{Example summaries on \hotel{} (first two) and \yelp{} (last four). Input opinions were filtered by the aspect categories (Asp), sentiment polarity (Pol), and \# of reviews (N). Colors show the alignments between opinions and summaries. Italic denotes incorrect extraction. Underlined opinions do not explicitly appear in the summaries.}\label{tab:controlled}
\vspace{-0.5cm}
\end{table*}

\subsection{Results}

\noindent\textbf{Automatic Evaluation:} Table~\ref{tab:results_yelp} shows the automatic evaluation scores for our model and 
the baselines on \yelp{} dataset. As shown, our framework outperforms all baseline approaches. Although \NAME{} is not a fully unsupervised framework, labeled data is only required by the opinion extractor and is easier to acquire than gold-standard summaries: on \yelp{} dataset, the opinion extraction models are trained on a publicly available ABSA dataset \citep{wang2017coupled}. 

\smallskip

\noindent \textbf{Human Evaluation:} We conducted three user studies
to evaluate the quality of the generated summaries (more details in Appendix~\ref{app:userstudy}).

First, we generated summaries from 3 systems (ours, LexRank and MeanSum/Best Review) for
every entity in \yelp's summarization test set and 200 random entities in the \hotel{} dataset, 
and asked judges to indicate the \textsl{best} and \textsl{worst} summary according to three criteria: \textsl{informativeness} (I), \textsl{coherence} (C), and \textsl{non-redundancy} (R). The systems' scores were computed using \textsl{Best-Worst Scaling} \citep{louviere2015best}, with values ranging from -100 (unanimously worst) to +100 (unanimously best.)
We aggregated users' responses and present the results in Table~\ref{tab:human_task1_all}(a). As shown, summaries generated by \NAME{} achieve the best informativeness and coherence scores compared to the baselines. However, \NAME{} may still generate redundant phrases in the summary. 

\looseness -1
Second, we performed a \textsl{summary content support} study. %
Judges were given 8 input reviews from \yelp{}, and a corresponding summary produced either by MeanSum or by our system. For each summary sentence, they were asked to evaluate the extent to which its content was supported by the input reviews. Table \ref{tab:human_task2} shows the proportion of summary sentences that were fully, partially, or not supported for each system. \NAME{} produced significantly more sentences with full or partial support, and fewer sentences without any support.

Finally, we evaluated our framework's ability to generate controllable output. We produced aspect-specific summaries using our \hotel{} dataset, and asked participants to judge if the summaries discussed the specified aspect exclusively, partially, or not at all. Table \ref{tab:human_task3} shows that in 46.6\% of the summaries exclusively summarized a specified aspect, while only 10.3\% of the summaries failed to contain the aspect completely.

\smallskip

\noindent {\bf Example Output:~} Example summaries in Table~\ref{tab:controlled} further demonstrate that a) \NAME{} is able to generate abstractive summaries from more than a hundred of reviews and b) produce controllable summaries by enabling opinion filtering.

The first two examples in Table~\ref{tab:controlled} show summaries that are generated from 8 and 128 reviews of the same hotel. \NAME{} performs robustly even for a large number of reviews. Since our framework is not based on aggregating review representations, the quality of generated text is not affected by the number of inputs and may result in better-informed summaries. This is a significant difference to previous work \cite{Chu:2019:MeanSum,Brainskas:2019:Unsupervised}, where averaging vectors of many reviews may hinder performance.

Finally, we provide qualitative analysis of the controllable summarization abilities of \NAME{}, which are enabled by input opinion filtering. As discussed in Section~\ref{sec:selection}, we filtered input opinions based on predicted aspect categories and sentiment polarity. The examples of {\sl controlled} summaries  (last 4 rows of Table \ref{tab:controlled}) show that \NAME{} can generate aspect/sentiment-specific summaries. 
These examples have redundant opinions and incorrect extractions in the input, but \NAME{} is able to convert the input opinions into natural summaries. Based on \NAME, we have built an online demo~\cite{ExtremeReader}\footnote{\url{http://extremereader.megagon.info/}} that allows users to customize the generated summary by specifying search terms.

\section{Conclusion}
We described \NAME{}, a simple yet powerful framework for abstractive opinion summarization. \NAME{} is a combination of existing ABSA and seq2seq models and does not require any gold-standard summaries for training. %
Our experiments on the \yelp{} dataset showed that \NAME{} outperforms baseline methods, including a state-of-the-art unsupervised abstractive summarization technique. Our user study and qualitative analysis
confirmed that our method can generate controllable high-quality summaries, and can summarize large numbers of input reviews.

\section*{Acknowledgements}
We thank Hayate Iso for helping debug the code. We also thank Prof. Mirella Lapata for helpful comments as well as the anonymous reviewers for their constructive feedback.

\bibliography{refs}

\begin{thebibliography}{22}
\expandafter\ifx\csname natexlab\endcsname\relax\def\natexlab#1{#1}\fi

\bibitem[{Amplayo and Lapata(2019)}]{amplayo2019informative}
Reinald~Kim Amplayo and Mirella Lapata. 2019.
\newblock Informative and controllable opinion summarization.
\newblock \emph{arXiv preprint arXiv:1909.02322}.

\bibitem[{Angelidis and Lapata(2018)}]{Angelidis:2018:MATE}
Stefanos Angelidis and Mirella Lapata. 2018.
\newblock Summarizing opinions: Aspect extraction meets sentiment prediction
  and they are both weakly supervised.
\newblock In \emph{Proc. EMNLP}, pages 3675--3686.

\bibitem[{Bražinskas et~al.(2019)Bražinskas, Lapata, and
  Titov}]{Brainskas:2019:Unsupervised}
Arthur Bražinskas, Mirella Lapata, and Ivan Titov. 2019.
\newblock Unsupervised multi-document opinion summarization as copycat-review
  generation.
\newblock \emph{arXiv preprint arXiv:1911.02247}.

\bibitem[{Cheng and Lapata(2016)}]{cheng-lapata-2016-neural}
Jianpeng Cheng and Mirella Lapata. 2016.
\newblock Neural summarization by extracting sentences and words.
\newblock In \emph{Proc. ACL '16}, pages 484--494, Berlin, Germany. Association
  for Computational Linguistics.

\bibitem[{Chu and Liu(2019)}]{Chu:2019:MeanSum}
Eric Chu and Peter~J. Liu. 2019.
\newblock {MeanSum}: {A} neural model for unsupervised multi-document
  abstractive summarization.
\newblock In \emph{Proc. ICML '19}, pages 1223--1232.

\bibitem[{Dai and Song(2019)}]{dai2019neural}
Hongliang Dai and Yangqiu Song. 2019.
\newblock Neural aspect and opinion term extraction with mined rules as weak
  supervision.
\newblock In \emph{Proc. ACL '19}, pages 5268--5277.

\bibitem[{Erkan and Radev(2004)}]{Erkan:2004:LexRank}
G{\"{u}}nes Erkan and Dragomir~R. Radev. 2004.
\newblock {LexRank}: Graph-based lexical centrality as salience in text
  summarization.
\newblock \emph{J. Artif. Intell. Res.}, 22:457--479.

\bibitem[{Hu and Liu(2004{\natexlab{a}})}]{Hu:2004:CustomerReviews}
Minqing Hu and Bing Liu. 2004{\natexlab{a}}.
\newblock Mining and summarizing customer reviews.
\newblock In \emph{Proc. KDD '04}, pages 168--177.

\bibitem[{Hu and Liu(2004{\natexlab{b}})}]{hu2004mining}
Minqing Hu and Bing Liu. 2004{\natexlab{b}}.
\newblock Mining opinion features in customer reviews.
\newblock In \emph{Proc. AAAI}, volume~4, pages 755--760.

\bibitem[{Isonuma et~al.(2019)Isonuma, Mori, and
  Sakata}]{Isonuma:2019:Unsupervised}
Masaru Isonuma, Junichiro Mori, and Ichiro Sakata. 2019.
\newblock Unsupervised neural single-document summarization of reviews via
  learning latent discourse structure and its ranking.
\newblock In \emph{Proc. ACL '19}, pages 2142--2152.

\bibitem[{Kiritchenko and
  Mohammad(2016)}]{Kiritchenko-Mohammad:2016:BestWorstScalingIsGood}
Svetlana Kiritchenko and Saif~M. Mohammad. 2016.
\newblock Capturing reliable fine-grained sentiment associations by
  crowdsourcing and best{--}worst scaling.
\newblock In \emph{Proc. NAACL-HLT '16}, pages 811--817.

\bibitem[{Li et~al.(2019)Li, Feng, Li, Mumick, Halevy, Li, and
  Tan}]{Li:2019:Opine}
Yuliang Li, Aaron Feng, Jinfeng Li, Saran Mumick, Alon Halevy, Vivian Li, and
  Wang-Chiew Tan. 2019.
\newblock Subjective databases.
\newblock \emph{Proc. VLDB Endow.}, 12(11):1330--1343.

\bibitem[{Lin(2004)}]{Lin:2004:ROUGE}
Chin-Yew Lin. 2004.
\newblock {ROUGE}: A package for automatic evaluation of summaries.
\newblock In \emph{Proc. ACL Workshop on Text Summarization Branches Out},
  pages 74--81.

\bibitem[{Liu and Lapata(2019)}]{liu-lapata-2019-hierarchical}
Yang Liu and Mirella Lapata. 2019.
\newblock Hierarchical transformers for multi-document summarization.
\newblock In \emph{Proc. ACL '19}, pages 5070--5081.

\bibitem[{Louviere et~al.(2015)Louviere, Flynn, and Marley}]{louviere2015best}
Jordan~J Louviere, Terry~N Flynn, and Anthony Alfred~John Marley. 2015.
\newblock \emph{Best-worst scaling: Theory, methods and applications}.
\newblock Cambridge University Press.

\bibitem[{Luo et~al.(2019)Luo, Li, Liu, and Zhang}]{luo2019doer}
Huaishao Luo, Tianrui Li, Bing Liu, and Junbo Zhang. 2019.
\newblock {DOER}: Dual cross-shared {RNN} for aspect term-polarity
  co-extraction.
\newblock In \emph{Proc. ACL '19}, pages 591--601.

\bibitem[{Miao et~al.(2020)Miao, Li, Wang, and Tan}]{Miao:2020:Snippext}
Zhengjie Miao, Yuliang Li, Xiaolan Wang, and Wang-Chiew Tan. 2020.
\newblock {Snippext}: Semi-supervised opinion mining with augmented data.
\newblock In \emph{Proc. WWW '20}.

\bibitem[{Paulus et~al.(2018)Paulus, Xiong, and
  Socher}]{Paulus:2017:DeepReinforced}
Romain Paulus, Caiming Xiong, and Richard Socher. 2018.
\newblock A deep reinforced model for abstractive summarization.
\newblock In \emph{Proc. ICLR '18}.

\bibitem[{See et~al.(2017)See, Liu, and Manning}]{See:2017:PointerGenerator}
Abigail See, Peter~J. Liu, and Christopher~D. Manning. 2017.
\newblock {Get To The Point}: {S}ummarization with pointer-generator networks.
\newblock In \emph{Proc. ACL '17}, pages 1073--1083.

\bibitem[{Vaswani et~al.(2017)Vaswani, Shazeer, Parmar, Uszkoreit, Jones,
  Gomez, Kaiser, and Polosukhin}]{Vaswani:2017:Transformer}
Ashish Vaswani, Noam Shazeer, Niki Parmar, Jakob Uszkoreit, Llion Jones,
  Aidan~N Gomez, {\L}ukasz Kaiser, and Illia Polosukhin. 2017.
\newblock Attention is all you need.
\newblock In \emph{Proc. NIPS '17}, pages 5998--6008.

\bibitem[{Wang et~al.(2017)Wang, Pan, Dahlmeier, and Xiao}]{wang2017coupled}
Wenya Wang, Sinno~Jialin Pan, Daniel Dahlmeier, and Xiaokui Xiao. 2017.
\newblock Coupled multi-layer attentions for co-extraction of aspect and
  opinion terms.
\newblock In \emph{Proc. AAAI '17}.

\bibitem[{Wang et~al.(2020)Wang, Suhara, Nuno, Li, Li, Carmeli, Angelidis,
  Kandogan, and Tan}]{ExtremeReader}
Xiaolan Wang, Yoshihiko Suhara, Natalie Nuno, Yuliang Li, Jinfeng Li, Nofar
  Carmeli, Stefanos Angelidis, Eser Kandogan, and Wang-Chiew Tan. 2020.
\newblock {ExtremeReader}: An interactive explorer for customizable and
  explainable review summarization.
\newblock In \emph{Companion Proc. WWW '20}, page 176–180.

\end{thebibliography}
\bibliographystyle{acl_natbib}

\newpage

\appendix

\section{Hyper-parameter Sensitivity Analysis}\label{app:sensitivity}
We present \NAME's hyper-parameters and their default settings in Table~\ref{tab:hyper-parameters}. Among these hyper-parameters, we found that the performance of \NAME{} is relatively sensitive to the following hyper-parameters: top-$k$ opinion ($k$), merging threshold ($\theta$), and maximum token length ($L$).

\begin{figure*}[h]
    \centering
     \subfigure[ROUGE-1]{ \includegraphics[width=0.32\textwidth]{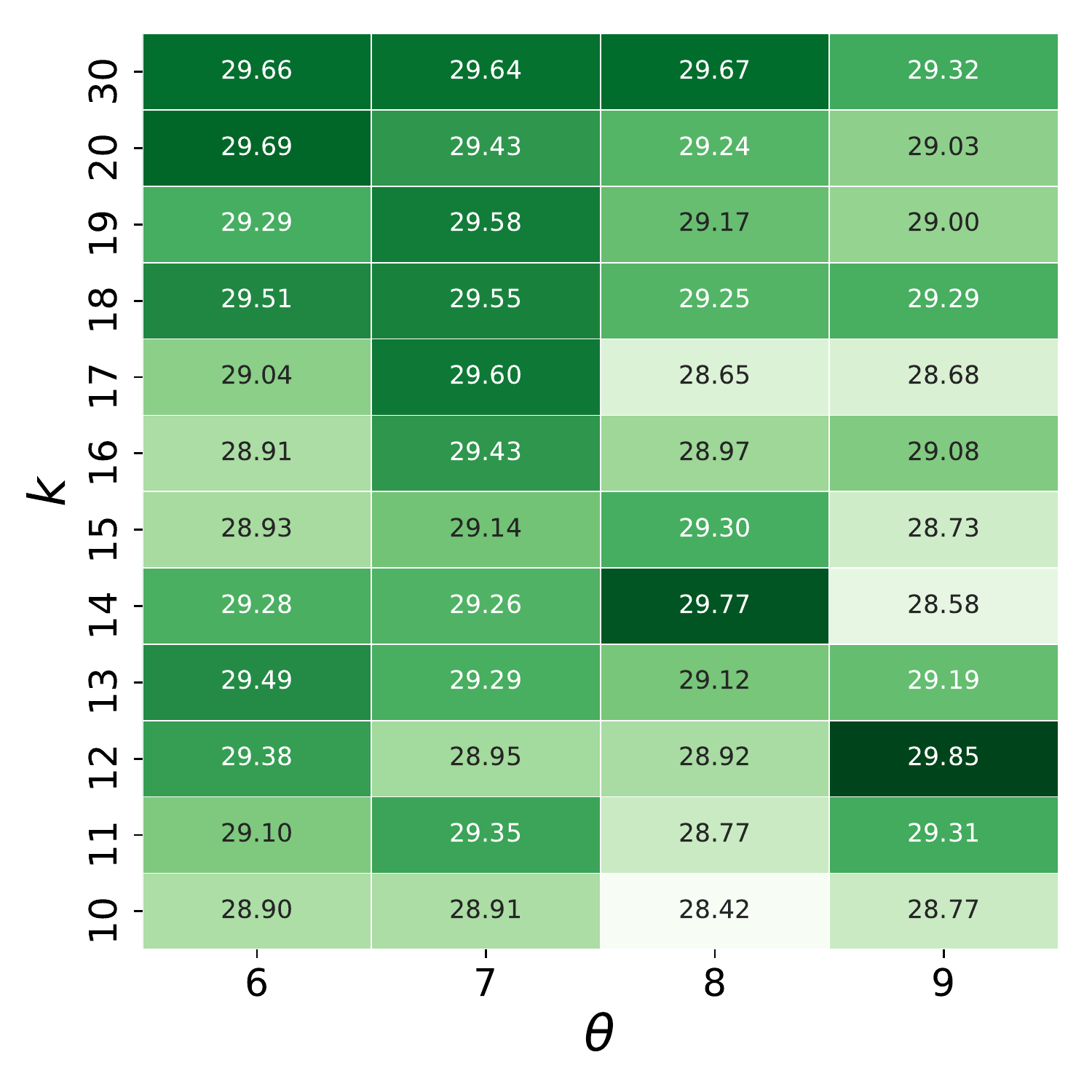}}
    \subfigure[ROUGE-2]{ \includegraphics[width=0.32\textwidth]{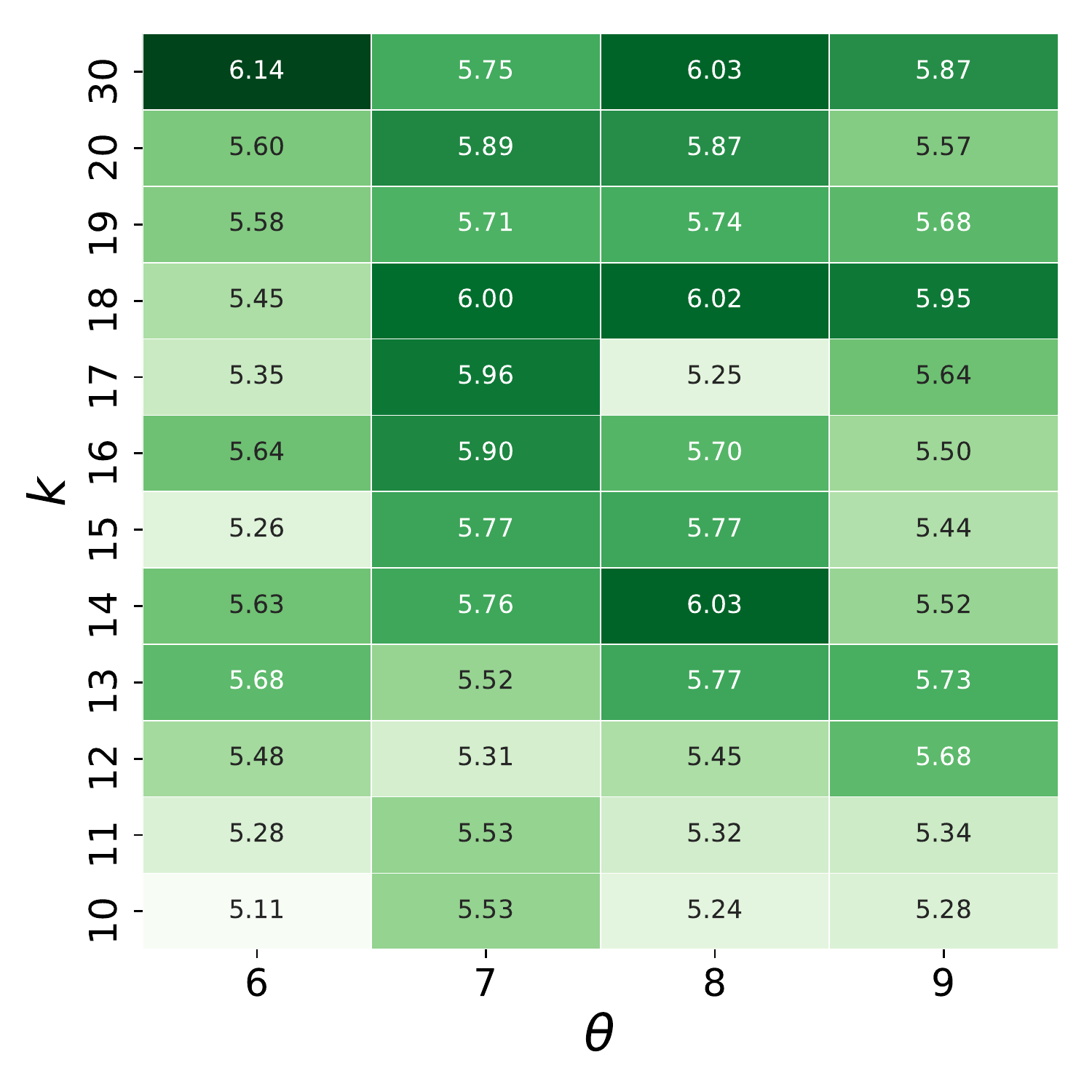}}
    \subfigure[ROUGE-L]{ \includegraphics[width=0.32\textwidth]{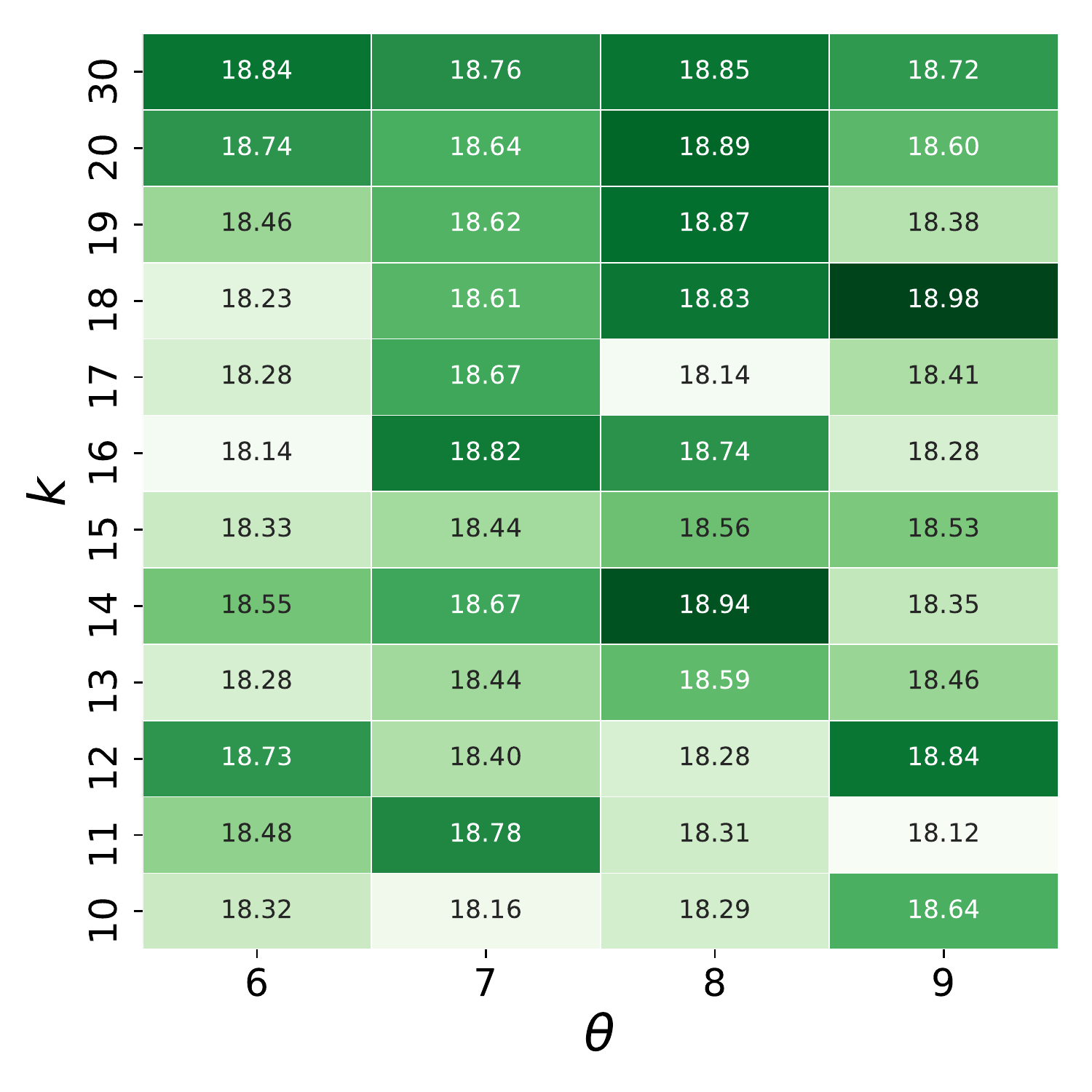}}    
     \subfigure[ROUGE-1]{ \includegraphics[width=0.32\textwidth]{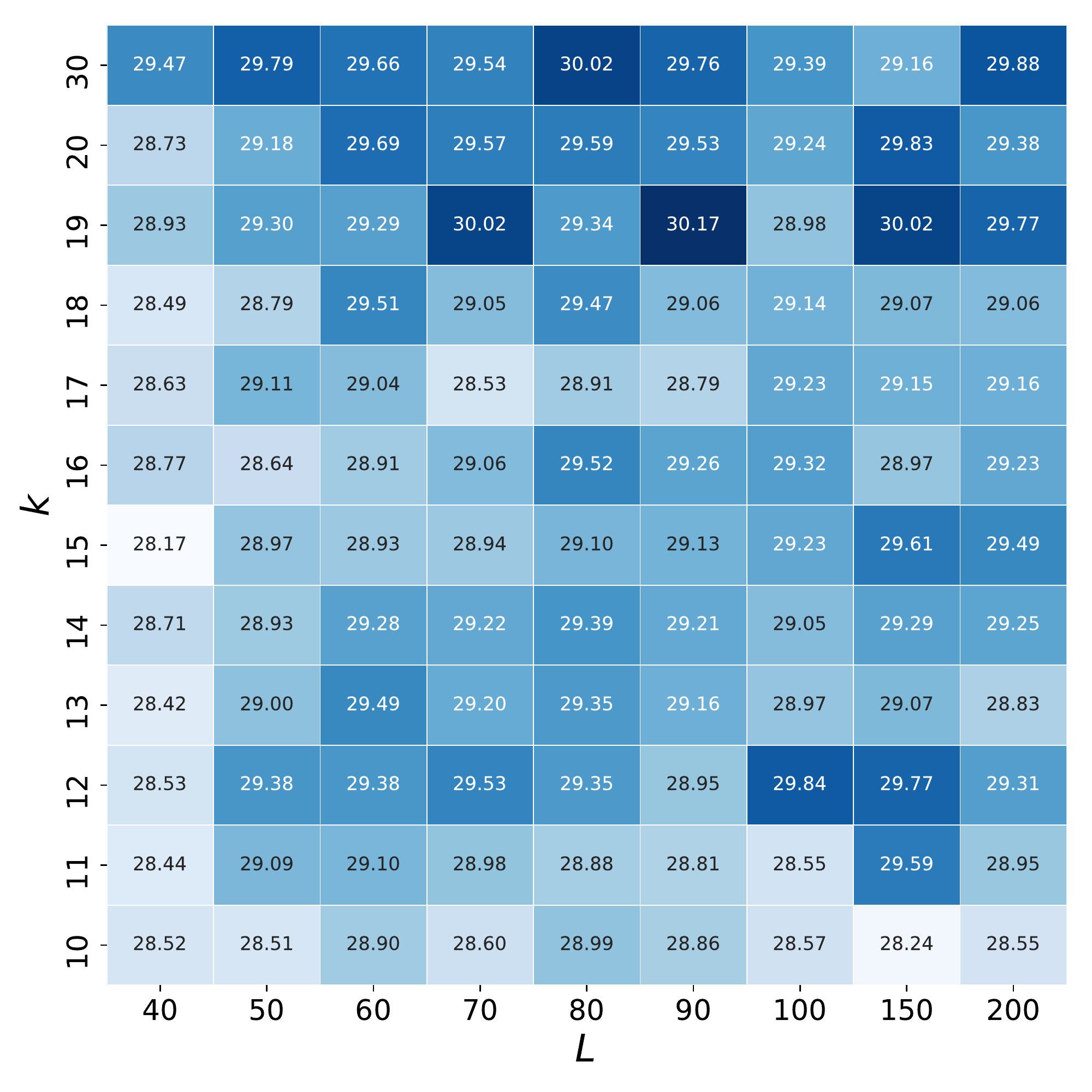}}
    \subfigure[ROUGE-2]{ \includegraphics[width=0.32\textwidth]{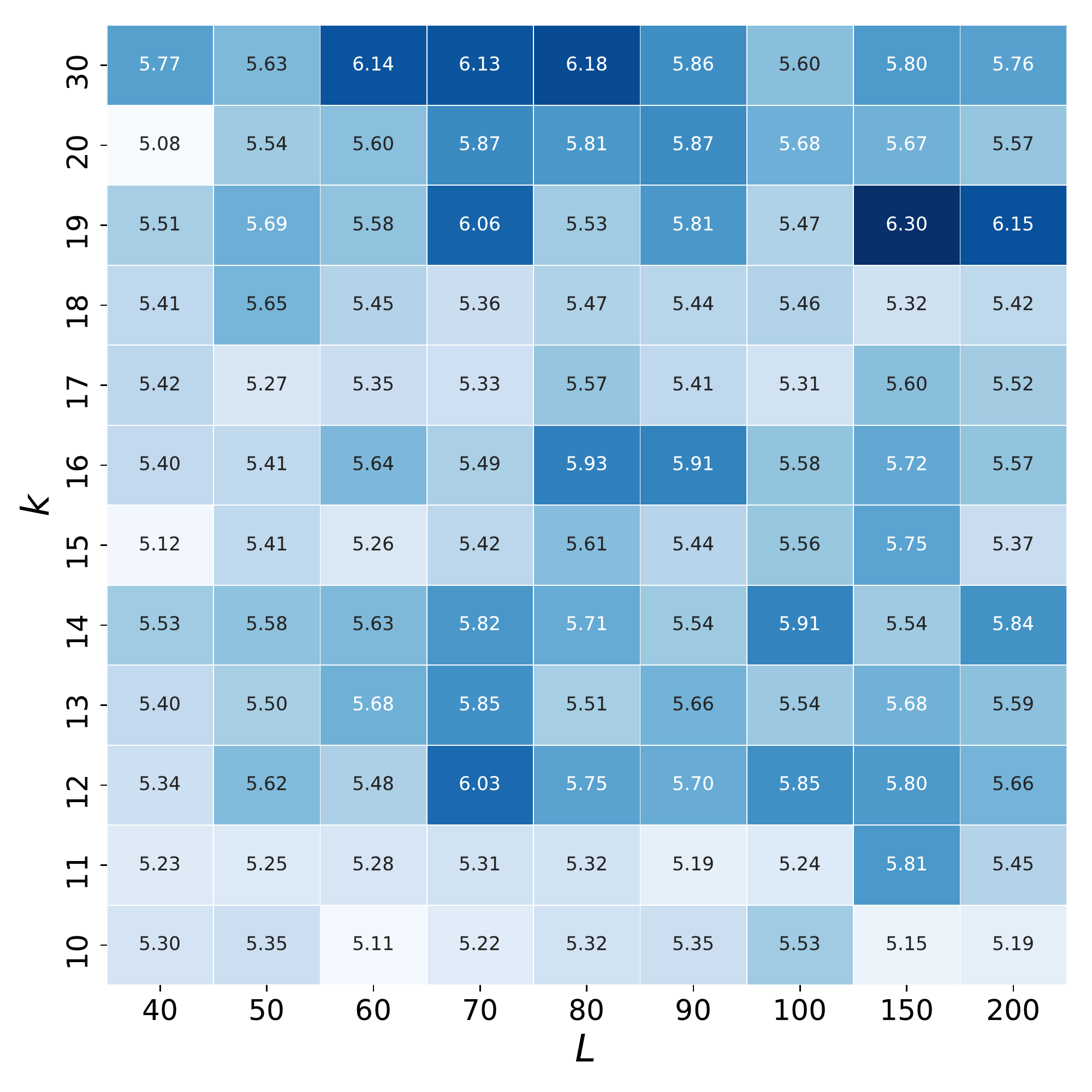}}
\subfigure[ROUGE-L]{ \includegraphics[width=0.32\textwidth]{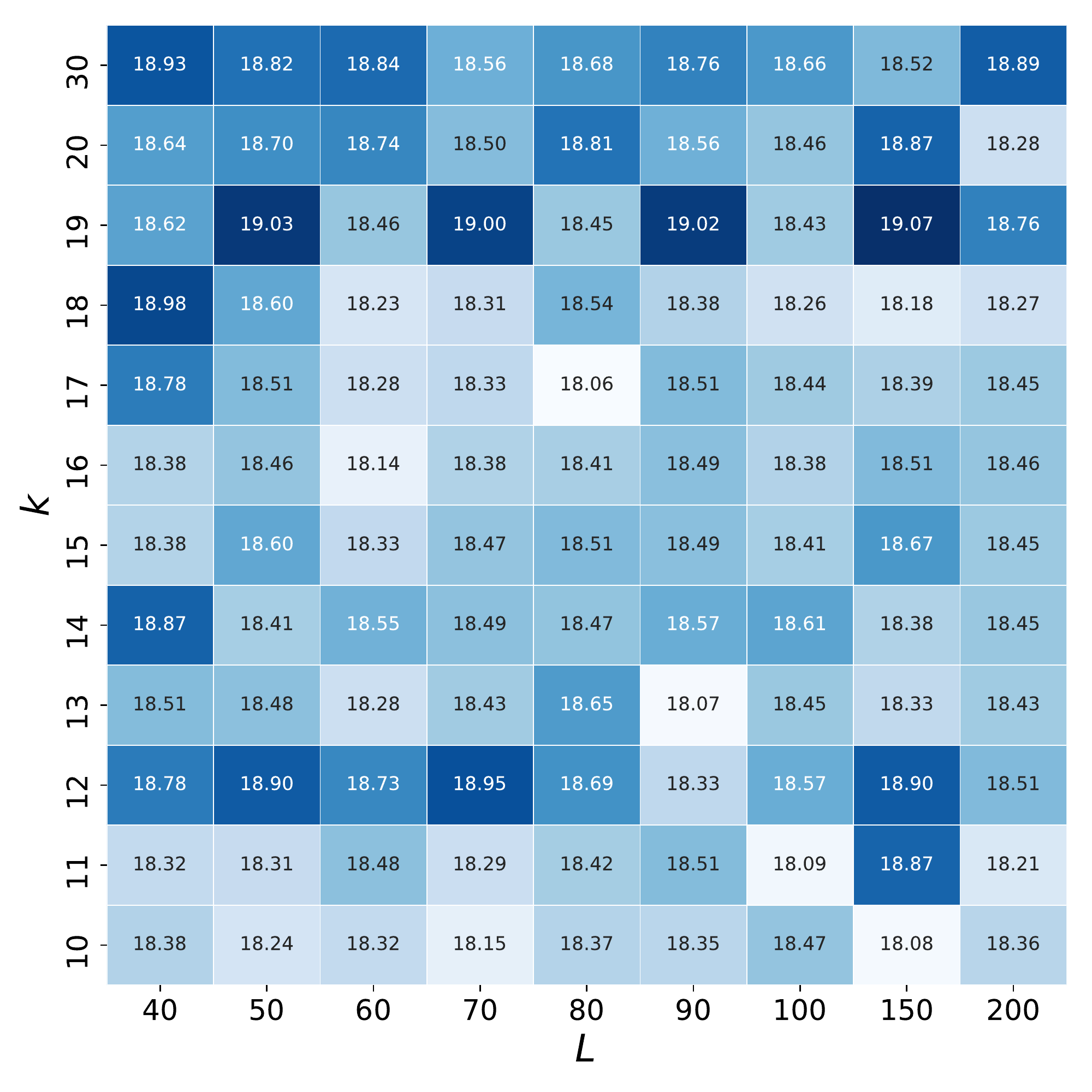}}
    \caption{Sensitivity analysis on hyper-parameters. Above row: Top-$k$ opinion ($k$) vs merging threshold ($\theta$); Bottom row: Top-$k$ opinion ($k$) vs max token size ($L$).}
    \label{fig:sensitivity}
\end{figure*}

To better understand \NAME's performance, we conducted additional sensitivity analysis of these three hyper-parameters. The results are shown in Figure~\ref{fig:sensitivity}. 

\noindent
{\bf Top-$k$ opinion vs Merging threshold:} We tested different $k=\{10, 11, \dots, 20, 30\}$ and $\theta=\{0.6, 0.7, 0.8, 0.9\}$. The mean (std) of R1, R2, and RL scores were 
29.2 ($\pm 0.3$), 5.6 ($\pm 0.2$), and 18.5 ($\pm 0.2$) 
respectively.

\noindent
{\bf Top-$k$ opinion vs Maximum token length:} We tested different $k=\{10, 11, \dots, 20, 30\}$ and $T=\{40, 50, \dots, 200\}$. The mean (std) of R1, R2, and RL scores were 
29.2 ($\pm 0.4$), 5.6 ($\pm 0.3$), and 18.5 ($\pm 0.2$) 
respectively.

The results demonstrate that \NAME{} is robust to the choice of the hyper-parameters and constantly outperforms the best-performing baseline method.

\section{Human Evaluation Setup}\label{app:userstudy}
We conducted user study via crowdsourcing using the FigureEight\footnote{\url{https://www.figure-eight.com/}} platform. To ensure the quality of annotators, we used a dedicated expert-worker pool provided by FigureEight. We present the detailed setup of our user studies as follows.

\begin{table}[t]
    \centering
    \begin{tabular}{lc}\hline
   \multicolumn{2}{l}{\textbf{Opinion Merging:}} \\\hline
   Word embedding & {\tt glove.6B.300d}\\
   Top-$k$ opinion ($k$) & 15\\
   Merging threshold ($\theta$) & 0.8 \\\hline
   \multicolumn{2}{l}{\textbf{Transformer model training:}} \\\hline
   SGD learning rate & 0.1  \\ 
   Momentum ($\beta$) & 0.1  \\
   Decay factor ($\gamma$) & 0.1 \\
   Number of epochs & 5  \\
   Training batch size & 8  \\\hline
   \multicolumn{2}{l}{\textbf{Decoding algorithm:}} \\ \hline
   Beam size & 5 \\
   Length penalty & 0.6 \\
   n-gram blocking ($n$) & 3 \\
   Maximum token length ($L$) & 60 \\\hline
    \end{tabular}
    \caption{List of \NAME{} hyper-parameters and the default settings.} 
    \label{tab:hyper-parameters}
    \normalsize
\end{table}

\begin{figure*}[h]
    \centering
    \includegraphics[width=\textwidth]{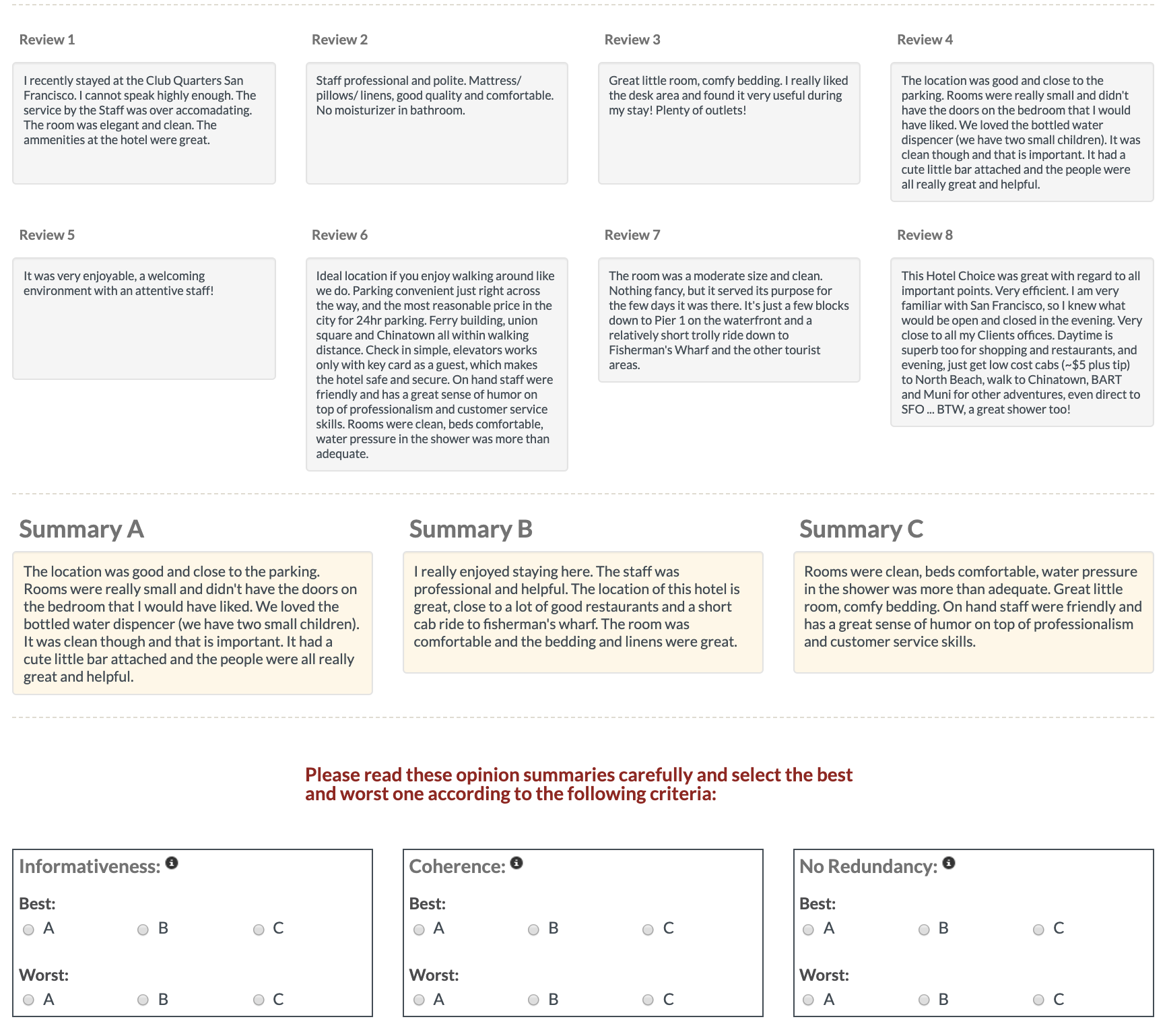}
    \caption{Screenshot of Best-Worst Scaling Task.}
    \label{fig:userstudy1}
\end{figure*}
\begin{figure*}[t]
    \centering
    \includegraphics[width=\textwidth]{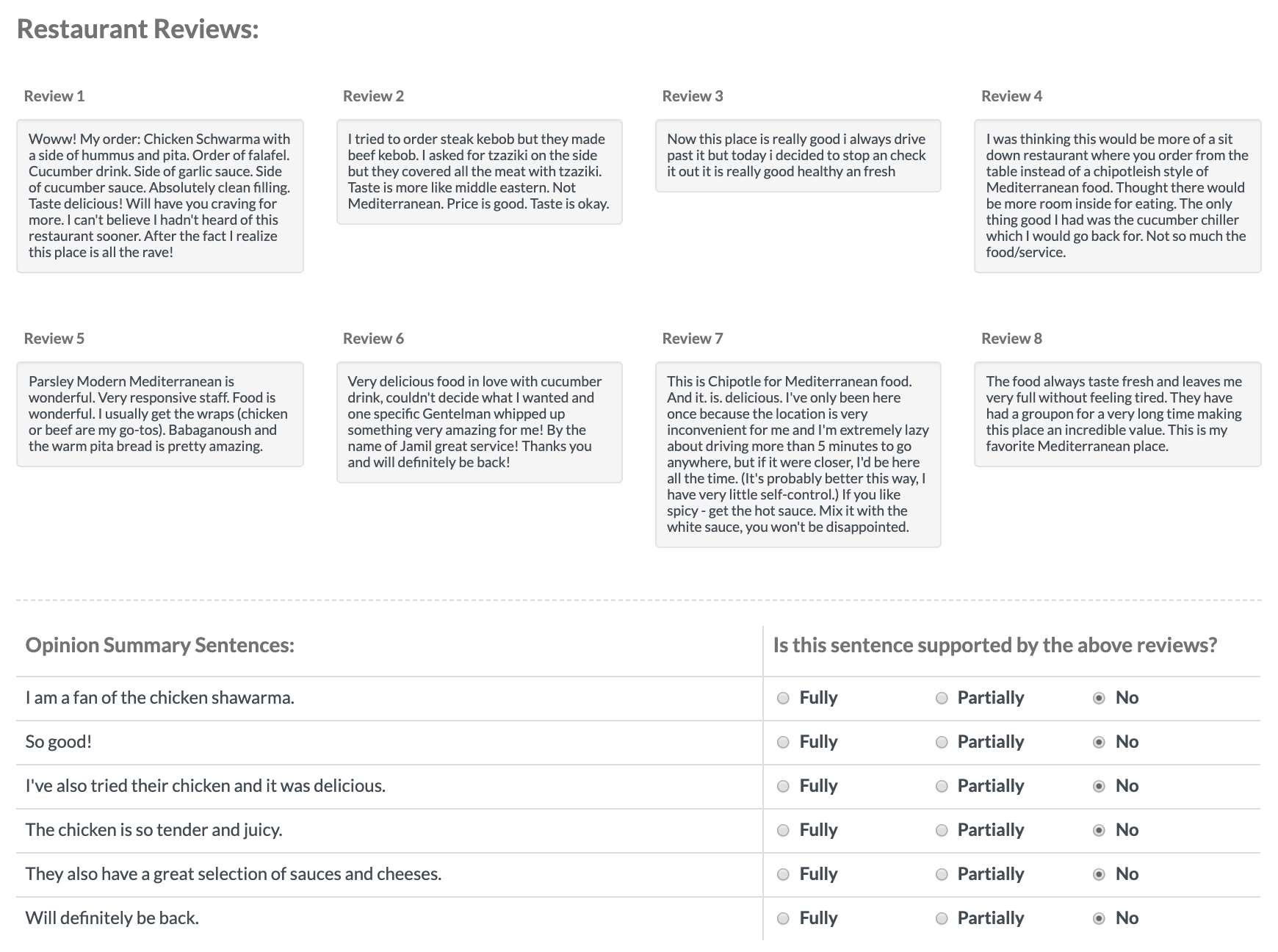}
    \caption{Screenshot of Content Support Task.}
    \label{fig:userstudy2}
\end{figure*}
\begin{figure*}[t]
    \centering
    \includegraphics[width=\textwidth]{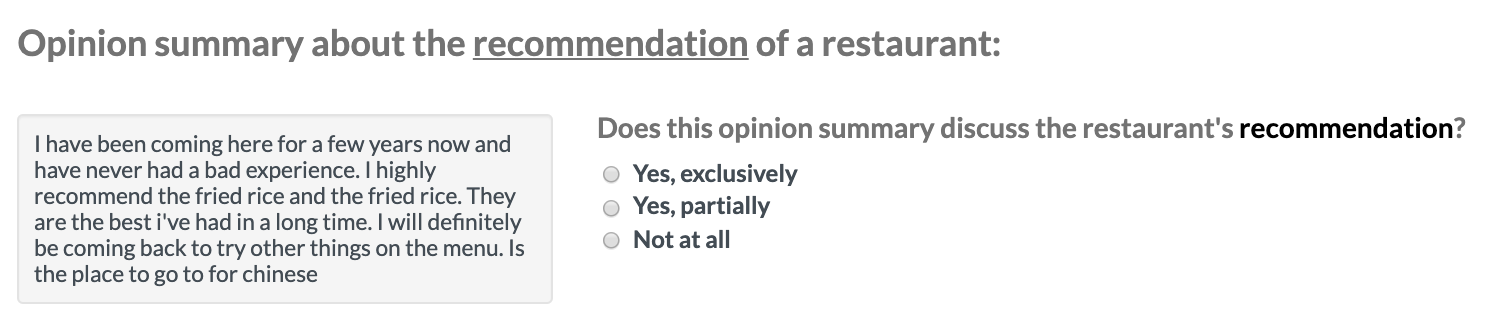}
    \caption{Screenshot of Aspect-Specific Summary Task.}
    \label{fig:userstudy3}
\end{figure*}

\paragraph{Best-Worst Scaling Task:} For each entity in the \yelp{} and \hotel{} datasets, 
we presented 8 input reviews and 3 automatically generated summaries to human annotators (Figure~\ref{fig:userstudy1}).  The methods that generated those summaries were hidden from the annotators and the order of the summaries were shuffled for every entity. We further asked the annotators to select the \textit{best} and \textit{worst} summaries w.r.t. the following criteria:

\begin{itemize}
    \item {\bf Informativeness:} How much useful information about the business does the summary provide? You need to skim through the original reviews to answer this.
    \item {\bf Coherence:} How coherent and easy to read is the summary?
    \item {\bf Non-redanduncy:} Is the summary successful at avoiding redundant and repeated opinions?
\end{itemize}
To evaluate the quality of the summaries for each criteria, we counted the number of best/worst votes for every system and computed the score as the \textit{Best-Worst Scaling}~\cite{louviere2015best}
:
\[
score = \frac{|\text{\it Vote}_{\text{\it best}}|-|\text{\it Vote}_{\text{\it worst}}|}{|\text{\it Votes}_{\text{\it all}}|}.
\]

The Best-Worst Scaling is known to be more robust for NLP annotation tasks and requires less annotations than rating-scale methods~\cite{Kiritchenko-Mohammad:2016:BestWorstScalingIsGood}.

We collected responses from 3 human annotators for each question and computed the scores w.r.t. informativeness (I-score), coherence (C-score), and non-redundancy (R-score) accordingly.

\paragraph{Content Support Task:} For the content support study, we presented the 8 input reviews to the annotators and an opinion summary produced from these reviews by one of the competing methods (ours or MeanSum). We asked the annotators to determine for every summary sentence, whether it is fully supported, partially supported, or not supported by the input reviews (Figure~\ref{fig:userstudy2}). We collected 3 responses per review sentence and calculated the ratio of responses for each category. 

\paragraph{Aspect-Specific Summary Task:} Finally, we studied the performance of \NAME{} in terms of its ability to generate controllable output. We presented the summaries to human judges and asked them to judge whether the summaries discussed the specific aspect exclusively, partially, or not at all (Figure~\ref{fig:userstudy3}). We again collected 3 responses per summary and calculated the percentage of responses.

\end{document}